\documentclass[conference]{IEEEtran}
\IEEEoverridecommandlockouts
\IEEEoverridecommandlockouts                              % This command is only
\usepackage{graphics} % for pdf, bitmapped graphics files
\usepackage{amsmath}
\usepackage{bm}

\DeclareRobustCommand{\uvec}[1]{{%
  \ifcsname uvec#1\endcsname
     \csname uvec#1\endcsname
   \else
    \bm{\hat{\mathbf{#1}}}%
   \fi
}}
\makeatletter
\newcommand*{\rom}[1]{\expandafter\@slowromancap\romannumeral #1@}
\makeatother
\usepackage{subcaption}
\usepackage{cite}
\usepackage{booktabs} % For formal tables
\usepackage{array}
\usepackage[english]{babel}
\usepackage[autostyle]{csquotes}
\usepackage[utf8]{inputenc}
\usepackage[english]{babel}
\usepackage{float}
\usepackage{amsmath,amssymb,amsfonts}
\usepackage{algorithmic}
\usepackage{graphicx}
\usepackage{textcomp}
\usepackage{xcolor}

\def\BibTeX{{\rm B\kern-.05em{\sc i\kern-.025em b}\kern-.08em
    T\kern-.1667em\lower.7ex\hbox{E}\kern-.125emX}}

\title{\LARGE \bf
Development of a Feeding Assistive Robot Using a Six Degree of Freedom Robotic Arm
}

\author{\IEEEauthorblockN{1\textsuperscript{st} Md Esharuzzaman Emu}
\IEEEauthorblockA{\textit{Dept. of Mechanical Engineering} \\
\textit{Northern Arizona University}\\
Flagstaff, AZ, USA \\
me692@nau.edu}
\and
\IEEEauthorblockN{2\textsuperscript{nd} Samarjith Biswas}
\IEEEauthorblockA{\textit{Dept. of Mechanical Engineering} \\
\textit{Northern Arizona University}\\
Flagstaff, AZ, USA\\
sb2679@nau.edu}
\and
\IEEEauthorblockN{3\textsuperscript{rd} Rajendra Shrestha}
\IEEEauthorblockA{\textit{Dept. of Mechanical Engineering} \\
\textit{Northern Arizona University}\\
Flagstaff, AZ, USA\\
rs2448@nau.edu}
}
\begin{document}

\maketitle
\thispagestyle{empty}
\pagestyle{empty}

%%%%%%%%%%%%%%%%%%%%%%%%%%%%%%%%%%%%%%%%%%%%%%%%%%%%%%%%%%%%%%%%%%%%%%%%%%%%%%%%
\begin{abstract}
This project presents the development of a Feeding Assistive Robot designed to aid individuals facing physical challenges, including those who have lost hand function or have limited arms control. The core component of the system is a 6-degree freedom robotic arm, which can be precisely controlled through voice commands. To implement this solution, a combination of readily available technologies, including an Arduino-based Braccio Arm, a distance sensor, and a Bluetooth module, were integrated to facilitate voice-controlled movements.

The primary objective of this assistive robot is to empower individuals with physical challenges to independently select and consume meals, whether at a dining table with family members or in the comfort of their beds. The Feeding Assistive Robot successfully achieves this goal, effectively assisting users in meal consumption through voice commands. The integration of a distance sensor and voice control enhances interactivity and responsiveness.

Notably, this assistive robot exhibits versatility by serving three different food items, catering to the user's preferences. This project represents a significant advancement in enhancing the quality of life for individuals with physical challenges, providing them with greater autonomy in daily activities.

\end{abstract}

%%%%%%%%%%%%%%%%%%%%%%%%%%%%%%%%%%%%%%%%%%%%%%%%%%%%%%%%%%%%%%%%%%%%%%%%%%%%%%%%
\section{INTRODUCTION}

The prevalence of disabilities in our global population is a matter of significant concern. According to the World Health Organization (WHO), approximately 15\% of the world's population, or nearly one in seven people, grapple with some form of disability, with 5.1\% experiencing difficulties related to mobility \cite{c1}. To put these numbers into perspective, in 2010, there were an estimated 785 to 975 million individuals under the age of 15 and older living with disabilities \cite{c2}. Moreover, aging populations are also susceptible to chronic illnesses that can lead to disabilities, with nearly 75\% of individuals aged 65 and over facing at least one chronic ailment, which can result in gradual or immediate loss of independence \cite{c3}.

Assistance and support are fundamental for individuals with disabilities to enhance their quality of life. Inadequate access to assistive support often forces disabled individuals to rely heavily on their families, creating a cycle of dependency that can hinder both the disabled person's autonomy and their family's economic participation. People with disabilities require support to enjoy a good quality of life and participate in social and economic activities on an equal footing with others \cite{c4}.

Researchers have been diligently working for decades to develop assistive devices that enable people with disabilities to feed themselves independently. Several have succeeded in designing automatic or semi-automatic feeding assistive devices, and a few are commercially available. These assistive devices encompass a range of options, including arm supports, human hand extensions for feeding, and electromechanical-powered devices \cite{c5}. Typically, such feeders consist of articulated, electrically powered arms with a spoon at the end, a plate mounted on a rotating turntable, and an auxiliary arm for food manipulation. Users control the movements of these components through switches, allowing for self-feeding \cite{c6}.

In the scope of this project, we took a cost-effective approach to empower individuals with disabilities through feeding assistance. Our aim was to repurpose a commercially available feeding arm, transforming it into an assistive feeding device that is more affordable than most alternatives on the market. The primary objective was to leverage a six-degree-of-freedom robotic arm to design a feeding assistive robot that emulates a natural feeding scenario, such as those experienced during daily dining. To achieve this objective, we meticulously configured the mechanical structure and workspace layout of the robotic arm on the dining table. We employed forward kinematics and an iterative method to precisely determine the position of the end effector. Furthermore, we ensured the safety and comfort of the robotic feeding arm within the workspace by incorporating a distance sensor and voice command functionalities.

In essence, this project represents a significant step towards enhancing the autonomy and quality of life for individuals with disabilities, particularly in the crucial aspect of mealtime independence. By repurposing existing technology in an innovative and cost-effective manner, we aim to contribute to a more inclusive and accessible future for those facing physical challenges.

\section{METHODS}
The development of the feeding assistive device involved the utilization of a commercially available 6-degree-of-freedom (6 DOF) Braccio robotic arm, coupled with the integration of a voice control module and a distance sensor. The system was meticulously engineered with the specific objective of predefining the positions of three distinct food items, ensuring a seamless and efficient feeding process. An essential assumption made during the design phase was that the user's mouth would maintain a consistent position throughout the feeding session.

The operation of the device relies on voice commands received from a mobile device, allowing the user to select their desired food item. Simultaneously, a highly responsive distance (ZX) sensor is employed to continuously gauge the distance between the user's mouth and the end effector of the robotic arm. This distance sensing capability is critical, as it ensures that the user has ample time to chew and swallow the food before the robotic arm repositions itself to retrieve the next chosen item from the user's predefined selection \cite{c7}.

To verify the precise positioning of the end effector, manipulator kinematics principles were harnessed. These calculations and validations serve as a crucial component of the system, guaranteeing that the robotic arm accurately reaches the intended location for each feeding action \cite{c8}. This level of precision is paramount to ensure the user's comfort, safety, and overall satisfaction during the feeding process.

\subsection{Mechanical System}
A braccio arm was used as the robotic arm for our experimental design. Although the Braccio arm has 6 degrees of freedom, only 5 degrees of freedom were used in our system. This is because the gripper rotation was fixed to attach the spoon and the spoon was considered as an end effector.\par
%fig 1
\begin{figure}[h]
\centering
    \begin{subfigure}{\linewidth}
    \centering
    \includegraphics[height=6cm, width=\linewidth]{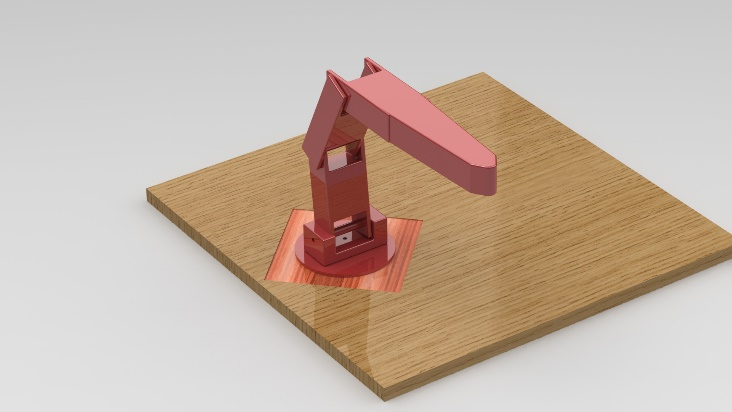}
    \caption{}
    \end{subfigure}
    \begin{subfigure}{\linewidth}
    \centering
    \includegraphics[height=6cm, width=\linewidth]{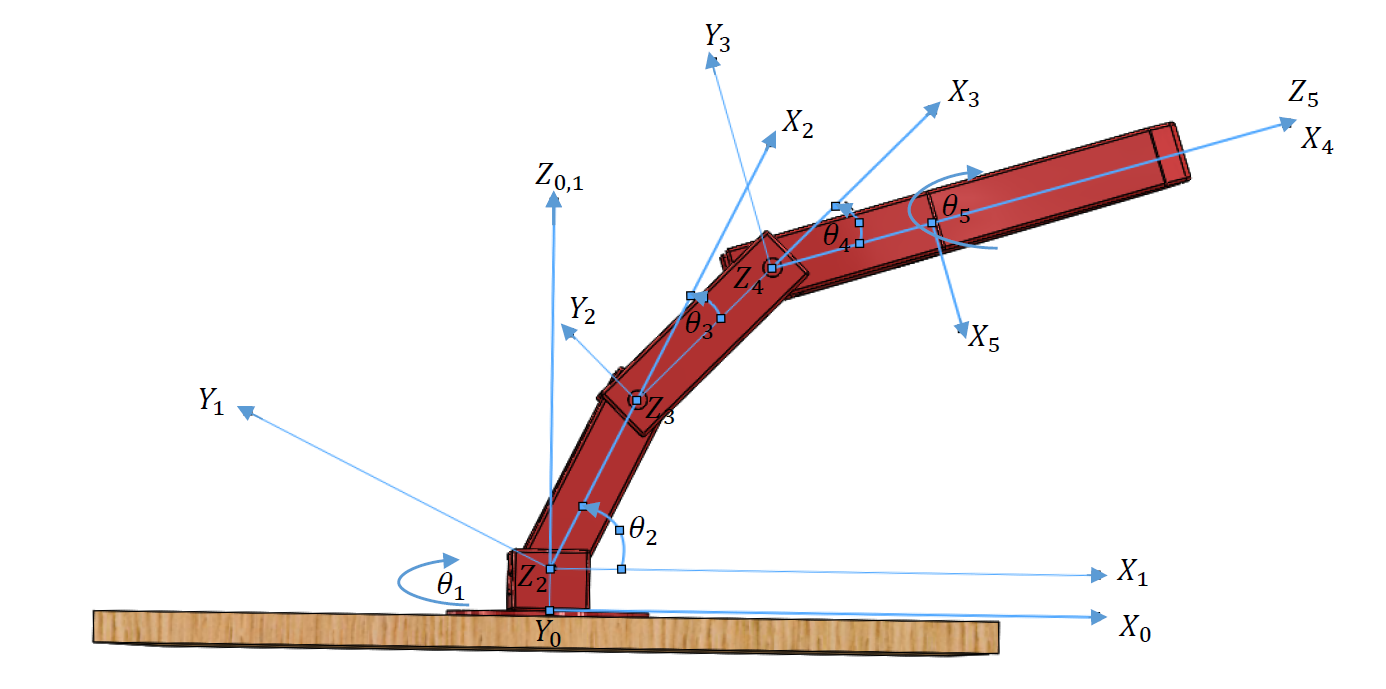}
    \caption{ }
    \end{subfigure}
\caption{(a)Schematic of simplified Braccio arm (modified 3D model), (b) Simplified model with the location of intermediate frames.}
\label{fig:my_label}    
\end{figure}

Figure 1(a) represents the simplified 3D model of a braccio arm which was modeled by SolidWorks. Here the wooden block was identified as the workspace of our system. This model was used to verify the forward kinematics of our robotic arm. Figure 1(b) represents the schematic of the braccio arm with all reference frames. This was used later to investigate the Denavit-Hartenberg Parameter\cite{c8}.

\subsection{Manipulator Kinematics}

In this paper forward kinematics have been used to find the end effector position in terms of the global coordinate\cite{c8}. Figure 2 represents the generalized link description for any kind of manipulator. Comparing figure 1(b) and figure 2, the D-H parameter of our robotic arm was listed in table \rom{2} where $a_i$ represents the distance from $\uvec{Z_i}$ to $\uvec{Z}_{i+1}$ measured along $\uvec{X_i}$, $\alpha_i$ represents the angle from $\uvec{Z_i}$ to $\uvec{Z}_{i+1}$ measured along $\uvec{X_i}$, $d_i$ represents the distance from $\uvec{X}_{i-1}$ to $\uvec{X_i}$ measured along $\uvec{Z_i}$ and $\theta_i$ represents the angle from $\uvec{X}_{i-1}$ to $\uvec{X_i}$ measured along $\uvec{Z_i}$.\par
%fig 2
\begin{figure}[ht]
\includegraphics[height=6cm, width=\linewidth]{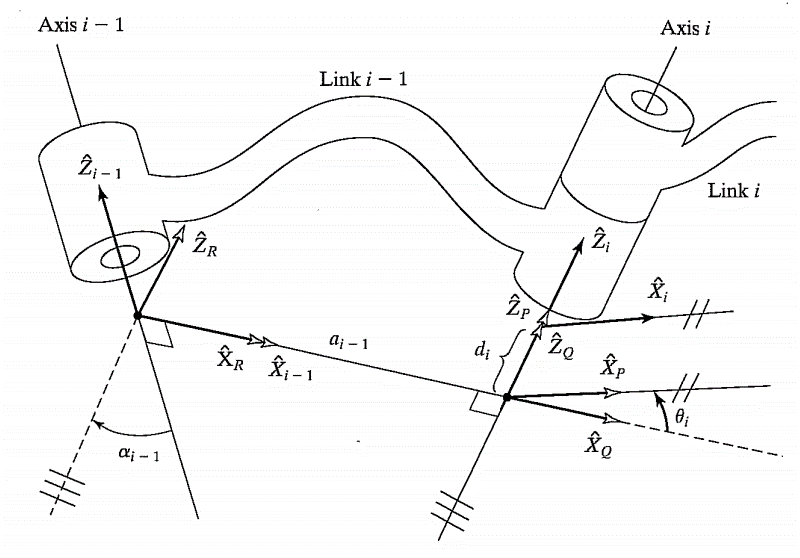}
\caption{Generalized link description and the location of intermediate frames \big\{P\big\}, \big\{Q\big\}, and \big\{R\big\}\cite{c8}.}
\label{fig:my_label}   
\end{figure}

%table 1
\begin{table}[H]
\caption{Length of different links of modified Braccio arm}
\label{table}
\begin{center}
\begin{tabular}{|c||c|}
\hline
Link Attributes & Length (Inch)\\
\hline
Base to Shoulder joint & 3\\
\hline
Shoulder joint Elbow joint & 5\\
\hline
Elbow joint to Wrist joint & 5\\
\hline
Wrist joint to End effector & 12\\
\hline
\end{tabular}
\end{center}
\end{table}

%table 2
\begin{table}[H]
\caption{Denavit-Hartenberg Parameter for Braccio arm}
\label{table}
\begin{center}
\begin{tabular}{|c||c||c||c||c|}
\hline
I & $\alpha_{i-1}$ & $a_{i-1}$ & $d_i$ & $\theta_i$\\
\hline
1 & 0 & 0 & 3 & $\theta_1$\\
\hline
2 & 90 & 0 & 0 & $\theta_2$\\
\hline
3 & 0 & 5 & 0 & $\theta_3$\\
\hline
4 & 0 & 5 & 0 & $\theta_4$\\
\hline
5 & 0 & 12 & 0 & $\theta_5$\\
\hline
\end{tabular}
\end{center}
\end{table}

The following general form of the equation has been used for the link transformation:
\begin{equation}\label{eqn}
{}^{i-1}_{i}T
=\begin{bmatrix}
c\theta_i & -s\theta_i & 0 & a_{i-1}\\
s\theta_{i}c\alpha_{i-1} & c\theta_{i}c\alpha_{i-1} & -s\alpha_{i-1} & -s\alpha_{i-1}d_i\\
s\theta_{i}s\alpha_{i-1} & c\theta_{i}s\alpha_{i-1} & c\alpha_{i-1} & c\alpha_{i-1}d_i\\
0 & 0 & 0 & 1
 \end{bmatrix}
 \end{equation}
 Using the equation \ref{eqn}, five transformation matrices $ {}^{0}_{1}T$, $ {}^{1}_{2}T$, $ {}^{2}_{3}T$, $ {}^{3}_{4}T$, $ {}^{5}_{4}T$ were found. The last column at each transformation matrix represents the position of the end effector with respect to the previous frame. Multiplying these five matrices, the final position of the end effector has been calculated with respect to the \{0\} frame.
\begin{center}
\begin{equation}\label{eqn}
{}^{0}_{5}T={}^{0}_{1}T \times {}^{1}_{2}T\times {}^{2}_{3}T \times  {}^{3}_{4}T \times {}^{4}_{5}T
 \end{equation}
\end{center}

\subsection{Hardware Design}
The total system includes a Braccio Robot Arm and motor shield, Arduino UNO, blue tooth module, ZX sensor, three bowls for food, and a wooden block as a dining place. Figure 1 shows the overview of the entire system over a dining table.
%fig 3
\begin{figure}
\centering
\includegraphics[height=7cm, width=\linewidth]{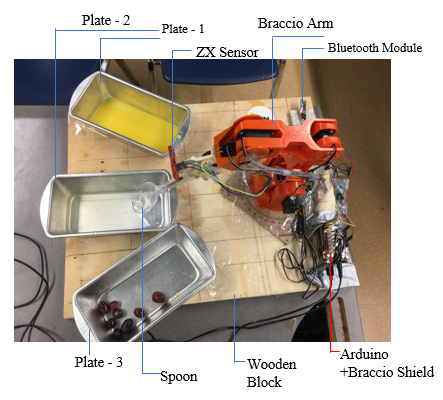}
\caption{Experimental setup of the robot system.}
\label{fig:my_label}   
\end{figure}
%fig 4
\begin{figure}
\centering
\includegraphics[height=7cm, width=\linewidth]{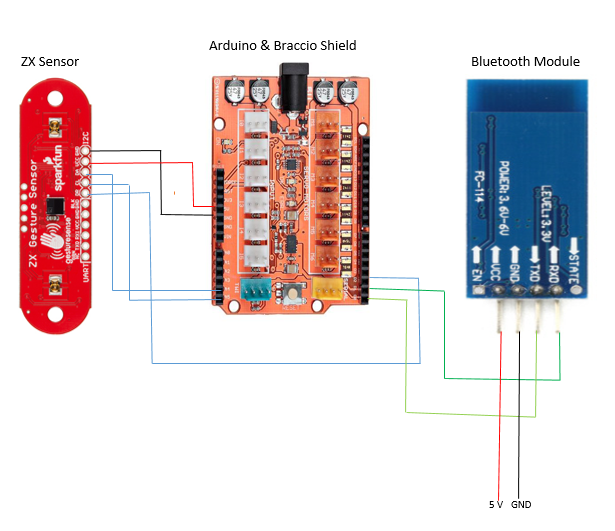}
\caption{Circuit diagram of a robot system.}
\label{fig:my_label}   
\end{figure}
%table 3
\begin{table}[H]
\caption{ZX sensor connection.}
\label{table}
\begin{center}
\begin{tabular}{|c||c|}
\hline
ZX Sensor & Arduino/Braccio Shield\\
\hline
VCC & 5V\\
\hline
GND & GND\\
\hline
DR & D2\\
\hline
CL & A5\\
\hline
DA & A4\\
\hline
\end{tabular}
\end{center}
\end{table}
%table 4
\begin{table}[H]
\caption{Bluetooth module connection.}
\label{table}
\begin{center}
\begin{tabular}{|c||c|}
\hline
Bluetooth Module & Arduino/Braccio\\
\hline
RX & TX\\
\hline
TX & RX\\
\hline
5V & VCC\\
\hline
GND & GND\\
\hline
\end{tabular}
\end{center}
\end{table}

\subsection{Controller Design}
The integration of the Bluetooth module (HC 06) plays a pivotal role in establishing seamless communication between the mobile device and the microcontroller, in this case, the Arduino Uno. Within this integrated framework, the mobile device's built-in microphone functions as a receiver for user voice commands, further leveraging the Google Voice recognition service to convert spoken words into text format.

The converted text, in the form of a string, is then transmitted to the Arduino input port via the Bluetooth device. Within the code developed for this purpose, the system meticulously scrutinizes the received text to determine if it corresponds to any of the predefined motion commands for the Braccio robotic arm. Upon identifying a match between the received command (string) and the predefined motions, the system proceeds to send the requisite signal to the Braccio shield, which is responsible for orchestrating the intricate movements of the robotic arm.

To ensure the user's safety and convenience, the ZX sensor comes into play by constantly monitoring the presence of the individual. In the event that the sensor detects the user's presence, it promptly transmits a command to the Arduino, instructing it to maintain the current position of the robotic arm. This feature is a crucial element of the system's operation, prioritizing user safety and experience.

For a comprehensive understanding of the system's connections and interactions, Figure 4 illustrates the connection diagram encompassing the ZX sensor, Arduino, and Bluetooth module. Additionally, Tables 3 and 4 provide a concise summary of the pin connections, facilitating an overview of the hardware configuration. A visual representation of the system's operational workflow can be found in Figure 5.
%fig 5
\begin{figure}[h]
\centering
\includegraphics[height=5cm, width=\linewidth]{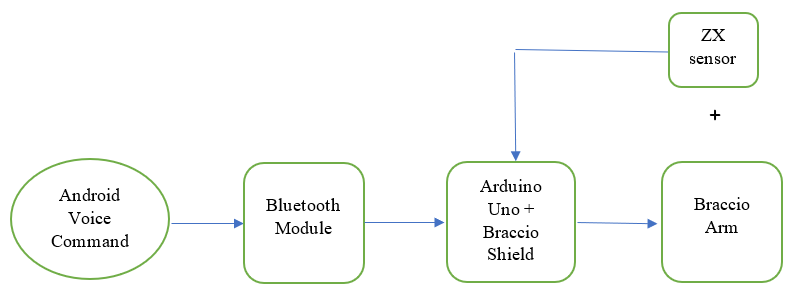}
\caption{Operation diagram of the robot system.}
\label{fig:my_label}   
\end{figure}
The motion of the arm was controlled by Arduino Uno as per the Arduino sketch (Appendix D) to achieve the desired goal of feeding a person\cite{c9}.

\subsection{Experimental Evaluation}
The robustness and functionality of the system were rigorously evaluated by conducting tests with a diverse group of individuals simulating the role of disabled users. These participants placed orders for various food items on a frequent basis, effectively assessing the system's ability to accommodate different user preferences. In these trials, the robot system successfully interpreted voice commands issued through the Android application and precisely executed the corresponding actions, facilitating the feeding process.

It is worth noting that the variability in pronunciation exhibited by different individuals for the same words presented a unique challenge. As a result, the Google Translate service occasionally encountered difficulties in accurately transcribing food names, leading to minor spelling errors. While this occurrence introduced a brief delay in the food ordering process, it is important to emphasize that all other aspects of the robot's performance remained consistently smooth and aligned with the desired outcomes.

The continuous operation of the ZX sensor, meanwhile, played a pivotal role throughout the testing phase. It continually collected data related to the user's positional information, providing invaluable input for decision-making within the system. Specifically, these real-time data inputs influenced decisions regarding whether to maintain the current arm position or proceed to the next designated food item. This dynamic feedback loop ensured that the robot system remained responsive to the user's needs and movements, contributing to an enhanced user experience and overall system efficiency.  
\section{Result}
To validate the positional accuracy of the Braccio arm's end effector, a MATLAB code was developed based on the Denavit-Hartenberg Parameters obtained from Table 2. The forward kinematics method was employed to calculate and visualize the end effector's position within the workspace over time, as depicted in Figure 6.

The analysis of Figure 6 revealed that the Braccio arm required approximately seven seconds to bring the food item close to the user's mouth. This feeding duration is fully customizable through adjustments to the servo motor speed, accommodating individual user preferences. In order to ensure smooth and safe arm movement, an iterative method was employed to determine the optimal angles for the Denavit-Hartenberg Parameters.

The system's responsiveness was a critical factor in its performance evaluation. On average, the system demonstrated a response time of 1.03 seconds for all movements, regardless of the specific command issued. Notably, the system exhibited a minimum response time of 0.93 seconds, both from the initial position and from any stable intermediate position. This rapid response time contributes significantly to creating a comfortable and secure feeding environment for the user.

Safety considerations were paramount in the system's design. In addition to an emergency power button, a "stop" command was incorporated to confirm the system's readiness for safe feeding conditions. This redundancy in safety measures ensures that the user's well-being remains a top priority.

Extensive testing was conducted, encompassing various food items and accommodating different user profiles and personalities. The system consistently demonstrated its reliability and effectiveness across all test cases, affirming its suitability for individuals with diverse needs and preferences. 
%fig 6
\begin{figure}[h]
\centering
    \begin{subfigure}{\linewidth}
    \centering
    \includegraphics[height=4cm, width=\linewidth]{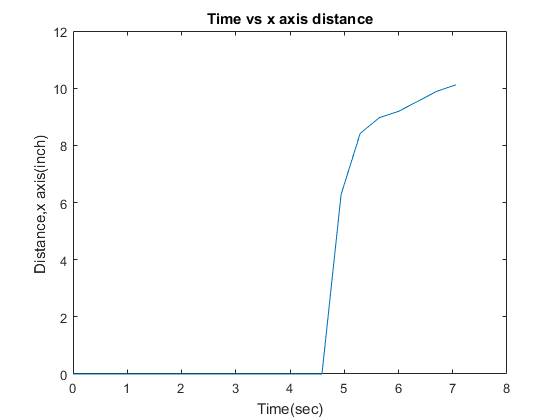}
    \caption{}
    \end{subfigure}
    \begin{subfigure}{\linewidth}
    \centering
    \includegraphics[height=4cm, width=\linewidth]{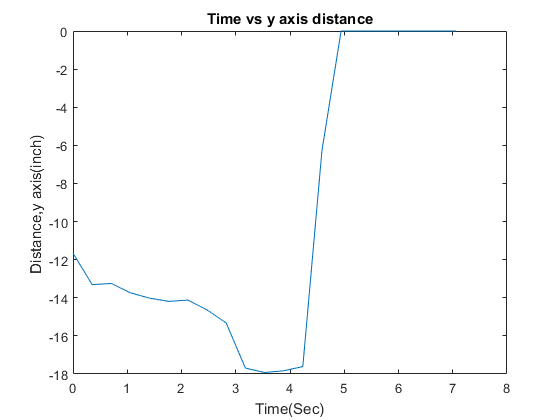}
    \caption{ }
    \end{subfigure}
    \begin{subfigure}{\linewidth}
    \centering
    \includegraphics[height=4cm, width=\linewidth]{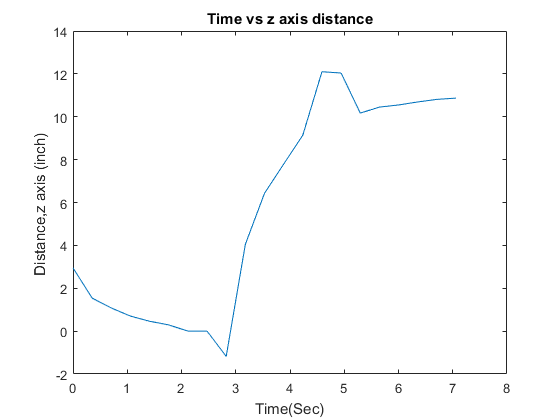}
    \caption{ }
    \end{subfigure}
\caption{End-effector position (a) Time vs X directional displacement. (b) Time vs Y directional displacement. (c) Time vs Z directional displacement.}
\label{fig:my_label}    
\end{figure}

\section{Discussion}
In the course of this project, we successfully achieved control of a commercially available 6-degree-of-freedom (6 DOF) Braccio robotic arm through voice commands and the integration of a distance sensor. This achievement represents a significant step forward in providing assistance to individuals with physical disabilities, particularly those who lack the use of an arm or have limited control over it. The Bluetooth module played a pivotal role in this process, acting as the conduit for transferring voice commands received from a mobile device to the Arduino microcontroller, which in turn orchestrated the precise manipulation of the robotic arm.

Validation of the desired positions for both the food items on the table and the user's mouth was accomplished through the application of forward kinematics. The ZX sensor continuously monitored the distance between the user's face and the robotic arm, ensuring that ample time was available to complete each step of the feeding process before the arm repositioned itself to retrieve the next ordered item.

A notable achievement of this project was the development of a feeding assistive robot at a cost of approximately $\$300$. This cost-effectiveness stands in stark contrast to the most prevalent commercially available feeding assistive device in the market, exemplified by the 'Ovi,' which can cost around $\$5590$. Additionally, our system's reliance on voice commands eliminates the need for touch-based controls, making it more accessible and flexible for individuals who may lack hand function.

To enhance the arm's interactivity and user-friendliness, we propose several potential additions. Integrating a voice receiver module within the arm would improve its responsiveness. Gesture and retina control mechanisms could make the device even more interactive and suitable for individuals who cannot or prefer not to speak. The inclusion of image processing capabilities would enable the arm to autonomously select specific foods based on user preference, eliminating the need for predefined food placement. Furthermore, modifying the end effector could expand the range of food options available to users, offering greater choice and personalization.

In conclusion, this project has demonstrated the feasibility of a cost-effective and user-friendly feeding assistive robot that holds great potential for improving the quality of life for individuals with disabilities. The proposed enhancements would further enhance the device's functionality and user experience, paving the way for more inclusive and adaptable assistive technology solutions.


\begin{thebibliography}{99}

\bibitem{c1} WHO, “World Report on Disability - Summary,” World Rep. Disabil. 2011, no. WHO/NMH/VIP/11.01, pp. 1–23, 2011.
\bibitem{c2} “World Health Organization, Evidence and Information for Policy,” World Heal. Surv. Geneva, World Heal. Organ. 2002–2004 ), 2002.
\bibitem{c3} T. J. H. Mrd-, B. Program, and L. P. Fried, “Disability in Older Adults: Evidence Regarding Significance, Etiology, and,” pp. 92–100, 1997.
\bibitem{c4} M. M. L. Verdonschot, L. P. De Witte, E. Reichrath, W. H. E. Buntinx, and L. M. G. Curfs, “Community participation of people with an intellectual disability: a review of empirical findings,” vol. 53, no. april, pp. 303–318, 2009.
\bibitem{c5} D. Mccoll and G. Nejat, “Meal - Time with a Socially Assistive Robot and Older Adults at a Long-term Care Facility,” vol. 2, no. 1, pp. 152–171, 2013.
\bibitem{c6} M. Guo, P. Shi, and H. Yu, “Development a Feeding Assistive Robot for Eating Assist,” pp. 299–304, 2017.
\bibitem{c7} Nhymel Shaw, “ZX Distance and Gesture Sensor Hookup Guide - learn.sparkfun.com.” [Online]. Available: https://learn.sparkfun.com/tutorials/zx-distance-and-gesture-sensor-hookup-guide. [Accessed: 08-May-2018].
\bibitem{c8} J. J. Craig, “Introduction to Robotics: Mechanics and Control 3rd,” Prentice Hall, vol. 1, no. 3, p. 408, 2004. 
\bibitem{c9} Guangming Xiong, Jianwei Gong, Taisen Zhuang, Tao Zhao, Dongxue Liu, Xijun Chen, "Development of Assistant Robot with Standing-up Devices for Paraplegic Patients and Elderly People", Complex Medical Engineering 2007. CME 2007, pp. 62-67, 2007. 
\end{thebibliography}
\end{document}